%% file: tmi.tex
\documentclass[journal,twoside,web]{ieeecolor}

\usepackage{etoolbox}
\makeatletter
\@ifundefined{color@begingroup}%
{\let\color@begingroup\relax
	\let\color@endgroup\relax}{}%
\def\fix@ieeecolor@hbox#1{%
	\hbox{\color@begingroup#1\color@endgroup}}
\patchcmd\@makecaption{\hbox}{\fix@ieeecolor@hbox}{}{\FAILED}
\patchcmd\@makecaption{\hbox}{\fix@ieeecolor@hbox}{}{\FAILED}

\usepackage{tmi}
\usepackage{cite}
\usepackage{booktabs}
\usepackage{multirow}
\usepackage{amsmath,amssymb,amsfonts}
\usepackage{hyperref}
\usepackage{algorithmic}
\usepackage{graphicx}
\usepackage{textcomp}
\usepackage{colortbl}
\usepackage{pifont}
\usepackage{subfigure}
\usepackage{float}
\definecolor{mygray1}{gray}{.9}
\definecolor{mygray2}{gray}{.7}
\newcommand{\cmark}{\ding{51}} 

\def\BibTeX{{\rm B\kern-.05em{\sc i\kern-.025em b}\kern-.08em
    T\kern-.1667em\lower.7ex\hbox{E}\kern-.125emX}}
\begin{document}
\title{HMIL: Hierarchical Multi-Instance Learning for Fine-Grained Whole Slide Image Classification}
\author{Cheng Jin, \IEEEmembership{Graduate Student Member, IEEE}, Luyang Luo, \IEEEmembership{Member, IEEE}, Huangjing Lin, Jun Hou, and Hao Chen, \IEEEmembership{Senior Member, IEEE}
\thanks{Received 13 December 2024; revised 14 December 2024; accepted 17 December 2024. Date of publication 20 December 2024; date of current version 3 April 2025. This work was supported by the National Natural Science Foundation of China (No. 62202403), Hong Kong Innovation and Technology Fund (Project No. MHP/002/22), Shenzhen Science and Technology Innovation Committee Fund (Project No. KCXFZ20230731094059008) and the General Program for Clinical Research at Peking University Shenzhen Hospital (No. LCYJ202001). \textit{(Corresponding author: Hao Chen.)}}
\thanks{Cheng Jin is with the Department of Computer Science and Engineering, The Hong Kong University of Science and Technology, Kowloon, Hong Kong SAR, China (e-mail: cheng.jin@connect.ust.hk).}
\thanks{Luyang Luo is with the Department of Computer Science and Engineering, The Hong Kong University of Science and Technology, Kowloon, Hong Kong SAR, China, and also with the Department of Biomedical Informatics, Harvard University, Cambridge, MA 02138 USA (e-mail: cseluyang@ust.hk).}
\thanks{Huangjing Lin is with Department of Computer Science and Engineering, The Chinese University of Hong Kong, Hong Kong, China (e-mail: hjlin@cse.cuhk.edu.hk).}
\thanks{Jun Hou is with the Department of Obstetrics and Gynecology, Peking University Shenzhen Hospital, Shenzhen 518036, China (e-mail: houjun0709@126.com).}
\thanks{Hao Chen is with the Department of Computer Science and Engineering and the Department of Chemical and Biological Engineering and Division of Life Science, The Hong Kong University of Science and Technology, Kowloon, Hong Kong SAR, China (e-mail: jhc@cse.ust.hk).}
\thanks{Digital Object Identifier 10.1109/TMI.2024.3520602}
}

\maketitle

\begin{abstract}
Fine-grained classification of whole slide images (WSIs) is essential in precision oncology, enabling precise cancer diagnosis and personalized treatment strategies. The core of this task involves distinguishing subtle morphological variations within the same broad category of gigapixel-resolution images, which presents a significant challenge. While the multi-instance learning (MIL) paradigm alleviates the computational burden of WSIs, existing MIL methods often overlook hierarchical label correlations, treating fine-grained classification as a flat multi-class classification task. To overcome these limitations, we introduce a novel hierarchical multi-instance learning (HMIL) framework. By facilitating on the hierarchical alignment of inherent relationships between different hierarchy of labels at instance and bag level, our approach provides a more structured and informative learning process. Specifically, HMIL incorporates a class-wise attention mechanism that aligns hierarchical information at both the instance and bag levels. Furthermore, we introduce supervised contrastive learning to enhance the discriminative capability for fine-grained classification and a curriculum-based dynamic weighting module to adaptively balance the hierarchical feature during training. Extensive experiments on our large-scale cytology cervical cancer (CCC) dataset and two public histology datasets, BRACS and PANDA, demonstrate the state-of-the-art class-wise and overall performance of our HMIL framework. Our source code is available at https://github.com/ChengJin-git/HMIL.
\end{abstract}


\input{Sections/1_Introduction}

\input{Sections/2_Related_Work}

\input{Sections/3_Method}

\input{Sections/4_Experiments}

\input{Sections/5_Conclusion}

\bibliographystyle{ieeetr}
\bibliography{reference}

\end{document}

%% file: Sections/1_Introduction.tex
\section{Introduction}
\label{sec:introduction} 
\IEEEPARstart{W}{hole-slide} images (WSIs) have been acknowledged as the gold standard for diagnosis \cite{siegel2022cancer, gurcan2009histopathological}. In precision oncology, fine-grained classification of WSIs is essential for accurate diagnosis and treatment planning. Unlike merely distinguishing between benign and malignant cases or simple categorization into two or three broad classes, fine-grained classification involves observing subtle morphological differences among cancer subtypes by examining different cell types and tissue structures within WSIs. This detailed classification provides doctors with more information to make accurate diagnoses and personalized treatment decisions, which is essential for recommending precise treatments such as surgery, radiation, and hormonal therapy \cite{elmore2015diagnostic}.

Significant challenges are presented in fine-grained WSI classification due to the need to differentiate subtle variations under the gigapixel resolutions inherent in WSIs, setting it apart from natural image classification tasks \cite{jin2023label}. To this end, multi-instance learning (MIL) has emerged as a leading approach for WSI classification. In this method, each slide is treated as a ``bag" containing multiple image patches (instances), and only the bag-level labels are required for training. Despite advancements in MIL, there has been limited progress in addressing fine-grained classification tasks within WSI. 

Hierarchical classification incorporates hierarchical labels and corresponding network designs to tackle fine-grained classification challenges \cite{silla2011survey, ran2023comprehensive}. In contrast to prior methods that address the problem in the setting of flat multi-class classification, hierarchical classification leverages the underlying structure of cancer subtypes. Several studies have attempted to address the challenges of fine-grained WSI classification within this context \cite{mercan2017multi, lin2021dual, gao2023childhood}. Specifically, Mercan et al. \cite{mercan2017multi} conceptualized this as a multi-instance, multi-label learning problem. They utilized a conventional max-pooling MIL method constrained by a multi-label loss, where the instances were regions of interest identified by pathologists. However, their approach did not incorporate the hierarchical mapping among cancer subtypes across different hierarchies, which has been empirically shown to enhance the performance of fine-grained image recognition in natural images \cite{nauta2021neural, chen2022label, zhou2023multi}. Introducing hierarchical mapping could provide valuable prior knowledge, aiding in distinguishing subtle differences between closely related subtypes.

Recognizing this potential, Lin et al. \cite{lin2021dual} proposed DPNet, which utilizes instance-level annotations along with a hierarchical grouping loss in the instance detector and a rule-based classifier for slide-level predictions. Gao et al. \cite{gao2023childhood} leverage information bottleneck theory to model pathologist-selected instances with hierarchical features within a multi-task framework, which employs an auxiliary instance-level classifier to enrich the feature representation for slide-level classification. While these approaches have advanced fine-grained WSI classification, their reliance on instance-level annotations limits broader applicability and fails to fully exploit hierarchical information for semantic guidance at both the instance and bag levels in MIL models.

\begin{figure*}
	\centering
	\includegraphics[width=\textwidth]{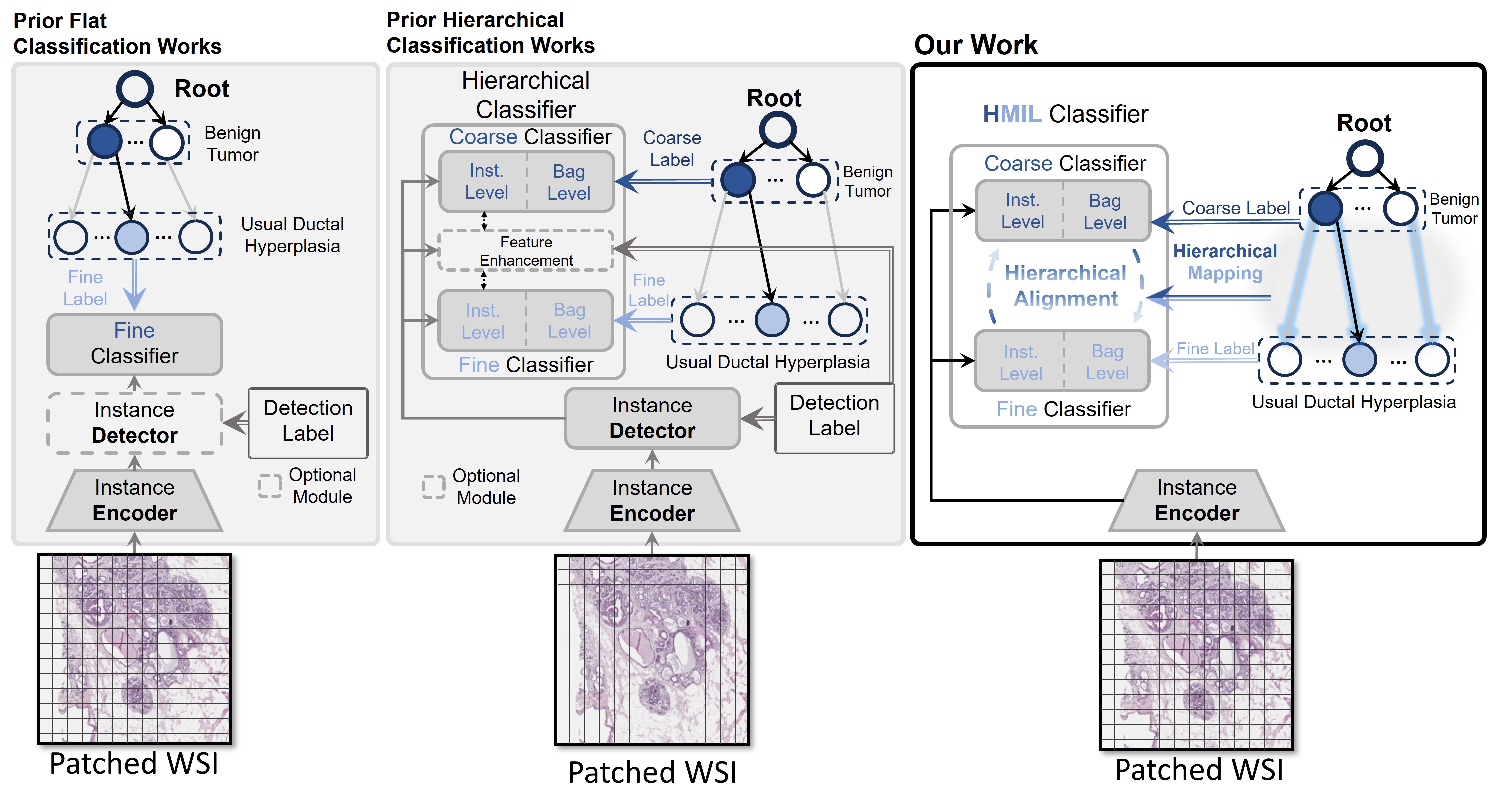}
	\caption{Comparison among prior works and our proposed HMIL framework in fine-grained WSI analysis. Left: Conventional flat classification methods, which form fine-grained classification as a multi-class classification task. Middle: Prior hierarchical classification methods, which typically leverage detector-enriched instance feature for hierarchical classification. Right: Our HMIL framework relaxed the need for detectors, introducing hierarchical alignment at both instance and bag level to improve fine-grained classification.}
	\label{fig:difference}
\end{figure*}

To this end, we propose a novel hierarchical multi-instance learning (HMIL) framework. As illustrated in Figure \ref{fig:difference}, our HMIL framework adopts a dual-branch structure: a coarse branch for coarse-grained classification and a fine branch for fine-grained classification. Between this dual-branch structure, we introduce hierarchical alignment at both instance and bag levels to better guide the learning process. At the instance level, both branches utilize class-wise attention-based MIL to introduce the foundation of hierarchical information, and the hierarchical instance matching module aligns the fine branch's class-wise attention with the coarse branch's class-wise attention through a fine-to-coarse similarity constrain. At the bag level, the hierarchical bag alignment module ensures fine-to-coarse prediction consistency by aligning the predictions of both branches. Moreover, we incorporate supervised contrastive learning \cite{khosla2020supervised} to strengthen the discriminative capability of the fine branch by maximizing inter-class distances and minimizing intra-class variations. Recognizing that the broad knowledge provided by the coarse branch may not sufficiently guide fine-grained classification, we introduce a dynamic weighting strategy to balance the influence between the coarse and fine branches during training.

The contributions of this paper are twofold. First, we formulate and explore hierarchical classification under the MIL settings and propose a novel framework termed HMIL. This framework leverages holistic hierarchical guidance at both the instance and bag levels to optimize the learning of feature embeddings and refine predictions, thereby enhancing the model’s ability to differentiate closely related cancer subtypes. Second, we evaluated our HMIL framework extensively on multiple fine-grained classification WSI datasets across various imaging modalities, including our private large-scale cytology cervical cancer (CCC) WSI classification dataset, which comprises 33,528 cytology WSIs, as well as two public histology WSI datasets, specifically BRACS \cite{brancati2022bracs} and PANDA \cite{bulten2022artificial}. Our findings indicate that HMIL achieves state-of-the-art performance compared to baseline models and enhances class-wise performance, revealing the importance of incorporating label hierarchy into the model.

%% file: Sections/2_Related_Work.tex
\section{Related Work}
\label{sec:related}
\subsection{MIL in WSI Classification}
In WSI classification, to tackle the challenges of gigapixel-sized WSIs with weak annotations, many methods primarily utilize the MIL framework. This framework involves three main stages: extracting features at the patch level, aggregating these patch-level features into slide-level representations, and training a classifier with these representations using slide labels for fully supervised prediction. Existing MIL methods in WSI classification can be broadly categorized based on their reliance on instance-level annotations.

Methods that rely on instance-level annotations typically leverage detailed region-specific information annotated by pathologists to enhance classification accuracy. Early works in histology WSI classification \cite{wang2016deep, bandi2018detection} and more recent works in cytology classification adopt this approach. These methods leverage patch-level annotations for training patch-based detection classification models that extract patch-level features, which are then aggregated to enable slide-level predictions within the MIL framework. For instance, Cheng et al. \cite{cheng2021robust} introduced a progressive detection method that utilizes multi-scale features for abnormal cell detection, followed by a recurrent neural network (RNN) \cite{zaremba2014recurrent} for slide-level classification. Cao et al. \cite{cao2021novel} enhanced detection performance by integrating clinical knowledge and an attention mechanism into their AttFPN cell detection model. Zhang et al. \cite{zhang2022whole} employed the RetinaNet \cite{lin2017focal} detection model for suspicious cell detection and the ResNeXt-50 \cite{hu2018squeeze} classification model for detection label refinement at instance level. At slide level, they aggregate them using graph attention networks (GAT) \cite{velivckovic2017graph} for WSI classification. However, these methods require labor-intensive, disease-specific manual annotations on the instances, limiting their applicability across different diseases.

In response, recent efforts have focused on developing frameworks that only require slide labels. Under this context, MIL methods can be further categorized into two directions: the design of feature extractors leveraging self-supervised methods and the exploration of various aggregation strategies \cite{bilal2023aggregation}. Advancements have been made in the pretraining of feature extractors \cite{li2021dual, chen2022scaling, yu2023prototypical, cao2023detection, zhao2024less} inspired by contrastive learning strategies in self-supervised learning \cite{chen2020improved, chen2020variational}. These methods aim to create robust feature representations that can be used in subsequent aggregation phase. At the aggregation phase, literature attempts to select the discriminative feature. Ilse et al. \cite{ilse2018attention} introduced aggregation based on instance-level attention scores, marking a seminar effort in this direction. Subsequently, Lu et al. \cite{lu2021data} developed a clustering-constrained attention MIL for WSI cancer classification, employing class-wise attention pooling to selectively emphasize on instances. Similarily, Zhang et al. \cite{zhang2023attentionchallenging} proposed multi-branch attention learning with stochastic masking strategy for discriminative instance discovery. Yu et al. \cite{yu2023prototypical} enhanced feature selection by extracting multiple cluster prototypes. From the perspective of alleviating the negative impact of insufficient training data, Zhang utilized bag augmentation by dividing training bags into smaller bags and applying double-tier feature distillation for training \cite{zhang2022dtfd}. Liu et al. employed a mixup approach for bag and label augmentation \cite{liu10385148}. Furthermore, innovative network architectures have been explored. Shao et al. employed the self-attention mechanism of the Transformer architecture \cite{vaswani2017attention} for histology WSI analysis, as exemplified by TransMIL \cite{shao2021transmil}. Recent advancements by Fillioux et al. \cite{fillioux2023structured} have investigated the structured state space model for long sequence modeling of patches within the MIL framework. Nevertheless, these techniques focus solely on a single resolution, which may neglect contextual nuances, prompting the development of multi-resolution methods \cite{li2021dual, chen2022scaling} to apprehend hierarchical features at different \textit{resolution levels}. 

These advancements underscore the potential of MIL models that require only slide-level labels compared to previous methods. However, existing methods primarily focus on binary or ternary classification tasks, which are relatively simple compared to fine-grained classification. 

\subsection{Hierarchical Fine-grained Recognition}
Fine-grained recognition is challenging due to small inter-class differences that complicate the distinction between similar categories. Conventional flat classifiers often fail to capture hierarchical relationships, limiting recognition accuracy. In response, hierarchical fine-grained recognition (HFR) assigns hierarchical labels to data points, enhancing the understanding of their relationships \cite{silla2011survey, ran2023comprehensive}. Typical HFR models follow a hierarchical architecture, with early designs featuring tree structures where leaf nodes represent specific classes and internal nodes indicate broader categories \cite{ji2020attention, chang2021your, kim2022vit}. Recent research leverages components such as knowledge graphs \cite{zhou2023multi} and hierarchical prompting \cite{wang2024transhp}, as well as strategies like self-paced learning \cite{yuan2024self}, to improve the capability of the model to learn hierarchical relationships.

Although these approaches demonstrate the potential of hierarchical classification in fully supervised learning contexts, a gap remains in applying such methods without relying on instance-level annotations in the MIL framework. For example, Mercan et al. \cite{mercan2017multi} employed a traditional max-pooling MIL approach constrained by a multi-label loss for breast cancer WSI classification, where the instances were regions of interest identified by pathologists. Lin et al. \cite{lin2021dual} explored cervical cancer screening on cytology WSIs using DPNet based on VGG-16 \cite{simonyan2014very}. A hierarchical grouping loss is proposed for suspicious cell detection, and the detected instances were aggregated with \textit{fixed} clinical rules at the bag level. Gao et al. \cite{gao2023childhood} proposed a multi-task framework for the classification of leukemia bone marrow. This framework utilizes the label hierarchy and introduces the information bottleneck achieved through contrastive methods on the instances \cite{tishby2000information, alemi2016deep}. Additionally, their approach leverages an instance-level auxiliary classifiers to enrich feature representation, aiming to improve classification accuracy. However, this method relies heavily on expert annotations, with each bag containing a relatively small number of pre-selected instances, which does not reflect the tens of thousands of instances typically involved in WSIs. Additionally, the neglect of alignment at the bag level restricts the capture of complex cellular features.

\begin{figure*}[h]
	\centering
	\includegraphics[width=\textwidth]{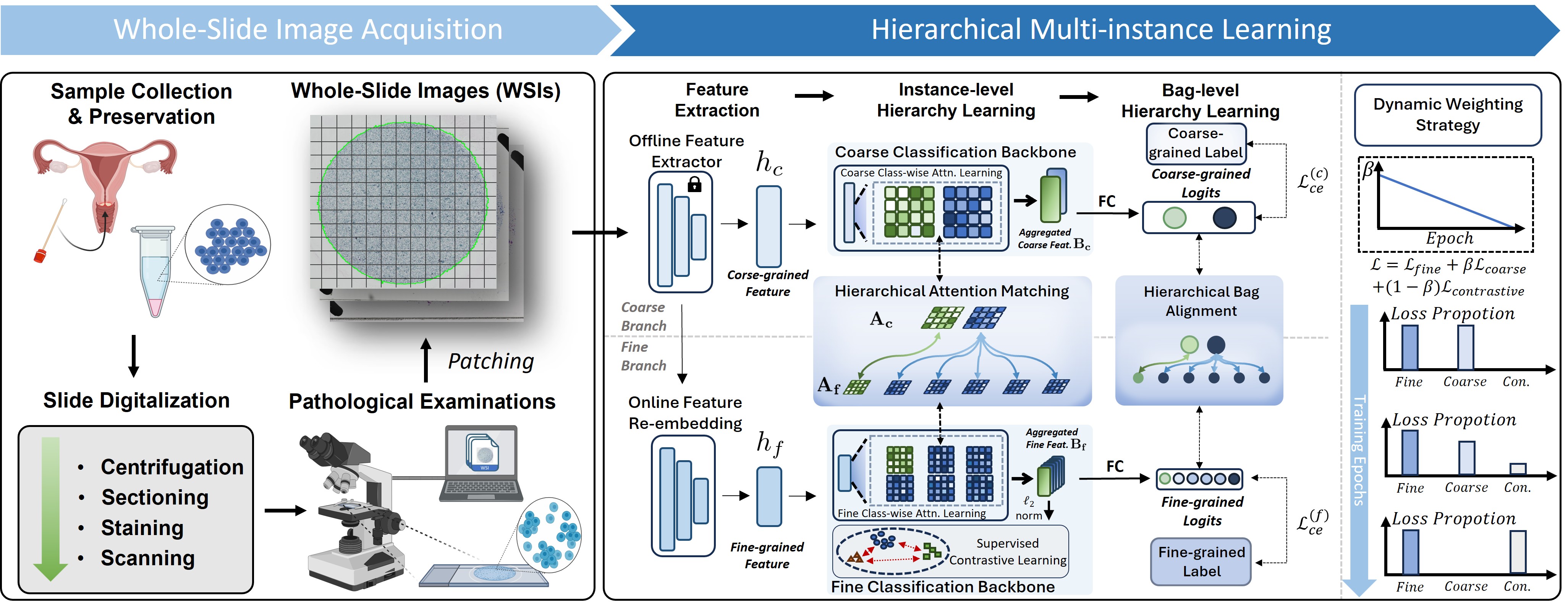}
	\caption{Overview of the proposed HMIL. We use fine-grained cervical cancer classification as an example. Patched WSI is fed into an offline feature extractor for the coarse features of the WSI, followed by an online feature re-embedding module that produces fine-grained feature. Subsequently, a dual-branch MIL architecture performs attention extraction and classification tasks at different hierarchical levels, with hierarchical alignment applied to instance and bag levels. Fully connected layers are then employed on top of the aggregated features in each branch to predict classification logits. Specifically, in the fine-grained branch, we incorporate supervised contrastive learning to further refine the feature representation. Finally, a dynamic weighting training strategy is incorporated to regulate the weights of these two branches throughout network training.}
	\label{fig:framework}
\end{figure*}

%% file: Sections/3_Method.tex
\section{Method}
\label{sec:method}

In this section, we first review the MIL paradigm and then highlight the distinctions of our method. We then introduce our HMIL framework, as illustrated in Figure \ref{fig:framework}.

\subsection{Preliminary}
\subsubsection{The MIL Paradigm}
From the perspective of MIL, a WSI $X$ is considered a bag, while its patches are considered instances within this bag, represented as $X = \{X_{i}\}_{i=1}^{N_i}$. The number of instances $N_i$ varies for different bags. For a binary classification task, there is a known label $Y$ for a bag and an unknown label $y_i$ for each of its instances. If there is at least one positive instance in a bag, then the bag is labeled as positive; otherwise, it is labeled as negative. The goal of a MIL model is to predict the bag label using all instances. As stated in the introduction, the MIL prediction process can be divided into three steps: instance feature extraction, aggregation, and bag classification, as follows:

\begin{equation}\label{eq1}
	\hat{Y} = h \left( g \left( \{ f(X_i) \}_{i=1}^{N_i} \right) \right)
\end{equation}

where $f(\cdot)$, $g(\cdot)$, and $h(\cdot)$ denote the instance feature extractor, aggregator, and bag classifier, respectively.

\subsubsection{Hierarchical Classification for MIL}
Hierarchical classification not only considers the presence of certain instances, but also leverages the predefined hierarchical mapping $\mathcal{M}(\cdot)$ reflecting the relationships between hierarchical labels to enhance classification performance. For a WSI bag $X$, its corresponding bag label $Y = (Y_{c}, Y_{f})$, where $Y_{c}=\mathcal{M}(Y_{f})$, represents the coarse-grained and fine-grained hierarchical labels, respectively. Each hierarchy contains $K_c$ and $K_f$ classes. Under this setting, existing methods \cite{lin2021dual, gao2023childhood} primarily leverage this mapping $\mathcal{M}$ at the instance level and requires instance annotation. In response, we advocate for using this mapping at both the instance and bag levels, while exploring the model's capability without instance annotations.

\subsection{The Hierarchical MIL Framework}
Our HMIL framework operates in a dual-branch hierarchical structure with a coarse branch and a fine branch. By leveraging label hierarchy comprehensively, we anticipate our framework not only learns from the broad categories provided by coarse labels but also effectively refines its predictions by focusing on the specific details and variations present in fine-level classes, enabling accurate fine-grained classification of WSIs.

\subsubsection{Hierarchical Feature Extraction}
Given a WSI $X$ tiled into $N_i$ instances, a pretrained encoder serves as an offline feature extractor (OFE), extracting coarse-grained feature vector $h_{c}$ with an embedded dimension of $d_c$. However, due to the differences in granularity required for specific classification tasks, it is necessary to re-embed these features. To address this, we leverage a non-linear multi-layer perceptron (MLP) serving as our online feature re-embedder (OFR) to re-embed these coarse-grained features into fine-grained feature vector $h_{f}$. The dimensionality reduction to $d_f = d_c/4$ is designed to refine the feature space and force the model to learn more discriminative features in a reduced feature space, thereby enhancing its ability to capture intricate patterns relevant to the classification task.

\subsubsection{Primal Hierarchical Guidance}
With the introduction of hierarchical labels, primal hierarchical guidance can be established by leveraging the classification loss. Based on these class-level probabilities, the objective functions for classification in the different hierarchies are defined using cross-entropy loss: $\mathcal{L}_{ce}^{(c,f)} = -\sum_{i=1}^{K_{c,f}} Y_i \log(\hat{Y}_i)$, where $Y$ is the true label, $\hat{Y}$ is the predicted probability distribution, and $K_{c,f}$ is the number of classes. The overall classification loss is then defined as: $\mathcal{L}_{cls}=\mathcal{L}^{(c)}_{ce}+\mathcal{L}^{(f)}_{ce}$. By applying this loss for coarse and fine classifications, we anticipate that it will provide foundational knowledge for the model to distinguish different subtypes of cancers. 

\subsubsection{Holistic Hierarchical Alignment}
Despite hierarchical classification losses provide basic guidance to the framework, relying solely on these losses overlooks the hierarchical relationships between categories. Without additional design considerations, priors from the coarse branch may introduce noise into fine-grained classification due to semantic misalignment. To address this, we introduce a holistic hierarchical alignment at both the instance and bag levels using a predefined hierarchical mapping matrix $\mathcal{M} \in \mathbb{R}^{K_f \times K_c}$, based on hierarchical relationships specified by pathologists. $\mathcal{M}$ maps fine categories to coarse categories, where each element $m_{i,j}$ is 1 if the fine-grained subtype $Y_f$ is a subtype of the coarse category $Y_c$, and 0 otherwise. This hierarchical alignment enables the model to semantically align features in the fine branch with those in the coarse branch, enhancing its ability to differentiate between nuanced cancer subtypes.

\textbf{Hierarchical Alignment at Instance Level.} At the instance level, our hierarchical alignment is achieved through the hierarchical attention matching (HAM) module. Recognizing the importance of hierarchical mapping, we employ class-wise attention learning instead of direct attention learning in our HAM module to effectively leverage the predefined mapping matrix $\mathcal{M}$. Specifically, we assess the contributions of the instances to the bag by utilizing the gated attention mechanism \cite{dauphin2017language} to learn the class-wise contributions of each instance within its respective hierarchy. The class-wise attention scores are computed as follows:
\begin{equation} 
\begin{aligned}
	\setlength\abovedisplayskip{0pt}
	\setlength\belowdisplayskip{0pt}
\mathbf{A_{\{c,f\}}} = \textrm{softmax}(&\mathbf{W_{\{c,f\}}}(\textrm{tanh}(\mathbf{V_{\{c1,f1\}}}(h_{\{c,f\}}))\\
&\odot \textrm{sigmoid}(\mathbf{V_{\{c2,f2\}}}(h_{\{c,f\}}))))
\end{aligned}
\end{equation}
here, $ \mathbf{A_{c}}\in\mathbb{R}^{K_c \times N_i} $ and $ \mathbf{A_{f}}\in\mathbb{R}^{K_f \times N_i} $ represent the instance attentions across classes at the coarse and fine levels, respectively. The fully connected (FC) layers $ \mathbf{V_{c1}} $, $ \mathbf{V_{c2}} $, $ \mathbf{W_{c}} $ and $ \mathbf{V_{f1}} $, $ \mathbf{V_{f2}} $, $ \mathbf{W_{f}} $ are designed with output dimensions of $ d_c/4 $, $ d_c/4 $, $ K_c $ and $ d_f/4 $, $ d_f/4 $, $ K_f $ respectively. This class-wise attention learning within each hierarchy allows the model to selectively emphasize more informative feature from the patches, enhancing the discriminative capability of the model across different levels of hierarchy. 

After obtaining class-wise attention at each hierarchy, we match the learned attention $\mathbf{A_{c}}, \mathbf{A_{f}}$ in our HAM module by aggregating the attention logits for corresponding classes as dictated by the mapping matrix $\mathcal{M}$. The alignment introduces an instance-specific coarse-to-fine constraint via a loss function defined as:
\begin{equation}
	\begin{aligned}
		\mathcal{L}_{ia} & = \frac{1}{N_i}(1-\cos(\mathbf{A_{i,c}}, \mathcal{M}\mathbf{A_{i,f}})) 
	\end{aligned}
\end{equation}
where $\cos$ denotes the cosine similarity, and the mapping matrix $\mathcal{M}$ translates fine-grained attention scores into the coarse-grained hierarchy. This strategic alignment ensure the fine-level learning does not deviate into incorrect or irrelevant feature spaces that do not align with the broader category defined at the coarse level. 

\textbf{Hierarchical Alignment at Bag Level.} At the bag level, alignment is centered on ensuring that predictions made at the fine level are meaningfully translatable back to the coarse level through the hierarchical bag alignment (HBA). We firstly obtain the prediction by utilizing attention pooling operations to aggregate class-wise instance-level features into bag-level representations, guided by the attention matrices $ \mathbf{A_{c}} $ and $ \mathbf{A_{f}} $: $\mathbf{B_{\{c,f\}}} = \,\, \mathbf{A_{\{c,f\}}}^{\top} \times h_{\{c,f\}}$, where $\times$ denotes matrix multiplication, $ \mathbf{B_{c}}\in\mathbb{R}^{K_c\times d_c} $ and $ \mathbf{B_{f}}\in\mathbb{R}^{K_f\times d_f}$ denote the bag-level representations at the coarse and fine levels, respectively. Subsequently, HMIL utilizes the bag-level representations from both hierarchy levels to compute the slide-level probabilities: $p_{\{c,f\}} = \text{softmax}(\mathbf{cls}_{\{c,f\}}(\mathbf{B_{\{c,f\}}}))$. In this formulation, $p_{c}$ and $p_{f}$ represent the probabilities that $X$ is classified into coarse and fine categories, respectively. These probabilities are determined by the classifiers $\mathbf{cls}_{c}$ and $\mathbf{cls}_{f}$, which consist of FC layers. The classifications for $X$ at both levels are obtained through $\hat{Y} = (\hat{Y}_{c}, \hat{Y}_{f}),$ where $\hat{Y}_{c} = \text{argmax}(p_{c})$ and $\hat{Y}_{f} = \text{argmax}(p_{f})$.

In HBA, given the fine-grained logits $p_{f}$, the mapping matrix $\mathcal{M}$ is employed to align the bag-level logits with their coarser counterparts can be expressed in a form analogous to the cross-entropy loss as follows:
\begin{equation}
	\mathcal{L}_{ba} = -\sum_{i=1}^{K_c} Y^{(c)}_i \log(\tilde{Y}^{(c)}_i),
\end{equation}
where $ Y^{(c)}_i $ is the true label for coarse category $ i $, and $ \tilde{Y}^{(c)} = \mathcal{M} p_{f} $ represents the predicted coarse probabilities derived from the fine probabilities through the mapping matrix $\mathcal{M}$. By enforcing the hierarchical alignment, the model is compelled to prevent the misinterpretation of fine-grained feature, and enhancing the overall accuracy of the classification.

\subsubsection{Supervised Contrastive Learning}
With the introduction of hierarchical alignment, given that fine-grained classification of WSI necessitates differentiating subtle variations inherent in gigapixel resolutions, which are characterized by high similarity between classes and significant intra-class variability. To further enhance the discriminative capability of the fine-grained bag-level feature, we apply supervised contrastive loss \cite{khosla2020supervised} in a batch $ b $ to the $\ell_2$-normalized fine-grained bag-level feature $ B_f $, as defined by the equation below:
\begin{equation}
	\mathcal{L}_{reg} =\sum_{i=1}^b-\frac{1}{\left|{P}_i\right|} \sum_{\mathbf{B_{p,f}} \in {P}_i} \log \frac{\exp \left(\mathbf{B_{i,f}} \cdot \mathbf{B_{p,f}}^{\top} / \tau\right)}{\sum_{\mathbf{B_{o,f}} \in {V}_i} \exp \left(\mathbf{B_{i,f}} \cdot \mathbf{B_{o,f}}^{\top} / \tau\right)}
\end{equation}
where ${V_i}=\left\{\mathbf{B_{i,f}}\right\}_{i \in[b]} \backslash\left\{\mathbf{B_{i,f}}\right\}$ denotes the set of current batch feature at the fine branch, excluding $\mathbf{B_{i,f}}$. Set ${P_i}=\left\{\mathbf{B_{j,f}} \in {V_i}: Y_{j,f}=Y_{i,f}\right\}$ comprises feature within the fine branch that share the same fine-grained label. The temperature hyperparameter $\tau$ is set to 0.1 following the literature \cite{chen2020improved, chen2020simple}, with ablation studies detailed in Sect. \ref{sec: ablation}. This constraint improves the discriminative ability of fine-grained features by bringing embeddings of the same class closer together and pushing those of different classes further apart.

\subsection{Training Strategy and Overall Loss Function}
To design the overall loss function, we recognize that the coarse branch's broad knowledge is insufficient for fine-grained classification due to differences in feature hierarchies. Inspired by \cite{wang2021contrastive, chen2023area}, which use dynamic weighting to balance loss components based on task relevance, we propose our dynamic weighting strategy. Initially, we emphasize coarse classification and alignment losses to improve fine-grained classification, as we believe the coarse classification task is inherently less complex. As training progresses, we shift our focus toward the fine branch's supervised contrastive learning to enhance feature representation in the fine branch as follows:
\begin{equation}
\begin{aligned}
    \mathcal{L} &= \beta \cdot (\mathcal{L}^{(c)}_{ce} + \mathcal{L}_{ia} + \mathcal{L}_{ba} ) + (1 - \beta) \cdot \mathcal{L}_{reg} + \mathcal{L}^{(f)}_{ce} 
\end{aligned}
\end{equation}
where $\beta = 1 - \frac{e}{E}$ is a dynamically adjusting weighting coefficient, with $E$ as the total number of epochs and $e$ as the current epoch. Further details on parameter ablation studies can be found in \ref{sec: ablation}.

%% file: Sections/4_Experiments.tex
\section{Experiments}
\label{sec:experiments}
\begin{figure*}[h]
	\centering
	\includegraphics[width=\textwidth]{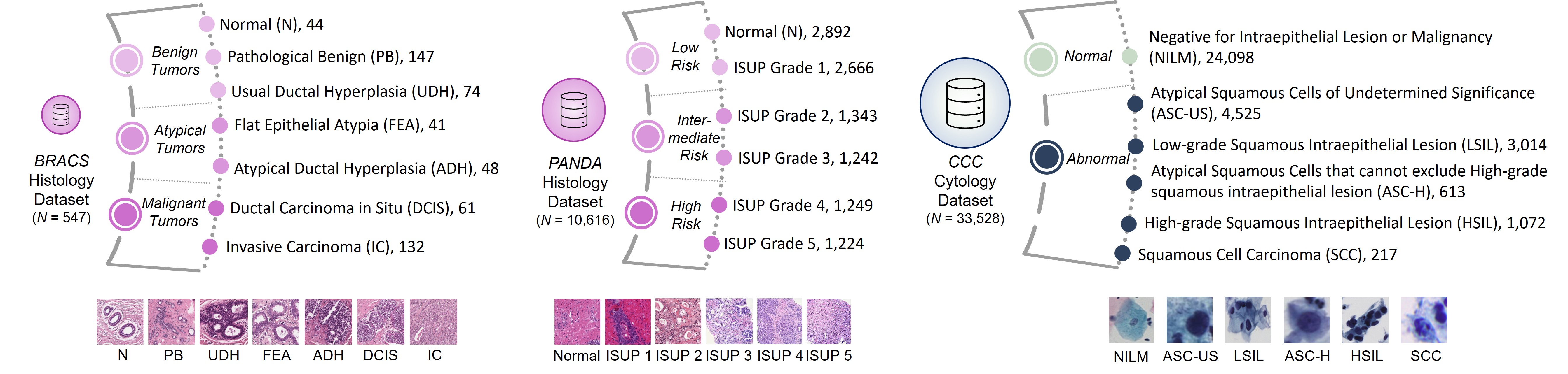}
	\caption{Hierarchical mappings and sub-class distributions in BRACS \cite{brancati2022bracs}, PANDA \cite{bulten2022artificial} and our collected CCC datasets. The mappings are from the original datasets designed by pathologists.}
	\label{fig:hierarchy}
\end{figure*}

\subsection{Datasets and Evaluation Metrics}
To assess the robustness and clinical applicability of our framework, we employed three datasets: two publicly available histology WSI datasets, namely BRACS \cite{brancati2022bracs} and PANDA \cite{bulten2022artificial}, along with our own collected cytology WSI dataset for cervical cancer screening, termed CCC. The details of the datasets are described as follows.

The \textbf{BReAst Carcinoma Subtyping (BRACS)} dataset is designed for breast cancer subtyping and comprises 547 histology WSIs. The dataset's labels are organized into a hierarchical structure to facilitate both coarse and fine-grained classification: at the coarse level, labels include benign tumors (BT), atypical tumors (AT), and malignant tumors (MT); at the fine level, labels are normal (N), pathological benign (PB), usual ductal hyperplasia (UDH), flat epithelial atypia (FEA), atypical ductal hyperplasia (ADH), ductal carcinoma in situ (DCIS), and invasive carcinoma (IC). Given the dataset's limited size, we employ a 10-fold cross-validation protocol.

The \textbf{Prostate cANcer graDe Assessment (PANDA)} dataset includes 10,616 histological WSIs of prostate biopsies, each annotated with a single label to indicate its normal status or corresponding ISUP  (International Society of Urological Pathology)  grade. Given the absence of the label hierarchy within the original dataset, we manually introduced coarse-level labels by mapping ISUP grades to risk categories as per the European Association of Urology (EAU) guidelines for prostate cancer \cite{salonia2023eau}. Specifically, WSIs categorized as normal or with an ISUP grade of 1 were assigned to the low-risk group. Those with an ISUP grade of 2-3 were classified as intermediate-risk, and biopsies with a grade of 4-5 were designated as high-risk. The original ISUP grades were retained as fine-level labels. A 10-fold cross-validation protocol was employed for both the training and testing phases.

Our in-house \textbf{Cervical Cytological Carcinoma (CCC)} dataset comprises 33,528 WSIs, collected from multiple medical centers. This dataset adheres to the Bethesda System (TBS) \cite{solomon2004bethesda} for cervical cytology classification, which delineates a range of cytological findings in the following hierarchical structure: labels include negative for intraepithelial lesion or malignancy (NILM) for specimens without cytological abnormalities, and five categories for positive findings: atypical squamous cells of undetermined significance (ASC-US), atypical squamous cells that cannot exclude high-grade squamous intraepithelial lesion (ASC-H), low-grade squamous intraepithelial lesion (LSIL), high-grade squamous intraepithelial lesion (HSIL), and squamous cell carcinoma (SCC). For benchmarking, the dataset was randomly divided into training, validation, and test sets at a ratio of 7:1:2 and employs non-parametric bootstrapping using 1,000 bootstrap replicates for testing to ensure the robustness of our evaluation.

The detail of the hierarchical mapping and sub-class distributions of these datasets can be referred to Figure \ref{fig:hierarchy}. To evaluate the classification performance of our datasets, we use a consistent set of metrics across different WSI classification tasks. Specifically, we report the metrics including accuracy, specificity, sensitivity, F1 score, and area under the curve (AUC) computed in a one-versus-rest manner. 

\subsection{Compared Baselines and Implementation Details}
We present the experimental results of our proposed HMIL framework compared to the following methods: (1) Conventional instance-level Multiple Instance Learning (MIL), which includes Mean-Pooling MIL and Max-Pooling MIL. (2) The standard attention-based MIL, ABMIL \cite{ilse2018attention}. (3) Four variants of ABMIL: the contrastive learning pretraining-based non-local attention pooling DSMIL \cite{li2021dual}, the single-attention-branch with clustering capability CLAM-SB \cite{lu2021data}, its multi-branch counterpart CLAM-MB \cite{lu2021data}, and the multi-branch attention-challenging ACMIL \cite{zhang2023attentionchallenging}. (4) Two transformer-based MIL architectures: TransMIL \cite{shao2021transmil} and the multi-resolution pretraining-based HIPT \cite{chen2022scaling}. (5) Pseudo bag augmented MIL, which includes double-tier augmented bag distillation DTFD \cite{zhang2022dtfd} and mixup-based bag augmentation PseMix \cite{liu10385148}. (6) Prototype-based metric learning MIL, PMIL \cite{yu2023prototypical}. (7) State space model-based MIL, S4MIL \cite{fillioux2023structured}. We faithfully reproduce these methods according to their official implementations.

During the preprocessing phase, we applied Otsu's thresholding method \cite{otsu1975threshold} to identify and delineate tissue regions for generating patches. Except for the DSMIL, HIPT, and PMIL methods, which used different patching strategies as specified in their original publications, we produced non-overlapping patches of $512 \times 512$ pixels at 20$\times$ magnification for the PANDA and BRACS datasets. For cytology WSIs, to accommodate varying resolutions across different imaging instruments, we standardized the images to a 20$\times$ magnification (0.2578 $\mu$m/pixel) and generated non-overlapping patches of $1,024 \times 1,024$ pixels. Following the studies in \cite{lu2021data,shao2021transmil,zhang2022dtfd}, we employed ResNet-50 \cite{he2016deep} as the offline feature extractor, except where DSMIL \cite{li2021dual}, HIPT \cite{chen2022scaling}, and PMIL \cite{yu2023prototypical} required different feature extractors according to their original papers. Specifically, DSMIL employs SimCLR \cite{chen2020simple} as the feature extractor and extracts features at 5$\times$ and 20$\times$ resolution with tiled patches of $224 \times 224$ pixels. HIPT employs the DINO \cite{caron2021emerging} approach and pretrains two vision transformer feature extractors at different resolutions, generating tiled patches of $256 \times 256$ and $4,096 \times 4,096$ pixels. We utilized the provided pretrained weights from the original work for the evaluation. PMIL finetunes ResNet-34 \cite{he2016deep} feature extractor using vocabulary-based prototype learning on the training split and generates tiled patches of $256 \times 256$ pixels.

The experiments were conducted on a workstation with NVIDIA RTX 3090 GPUs using the Adam optimizer with a weight decay of $1 \times 10^{-5}$. Training lasted for 200 epochs, during which the best results were saved. For the CCC and PANDA datasets, the learning rate was $1 \times 10^{-3}$ with a batch size of $b=512$. For the BRACS dataset, the learning rate was $1 \times 10^{-4}$ with a batch size of $b=128$.

\begin{table*}[h]
	\centering
	\caption{Evaluation of performance on the histology WSI datasets BRACS. We report the results in the form of $\text{Mean}_{\text{Std}}$. The best and second best results are highlighted in \textcolor{red}{red} and \textcolor{blue}{blue}, respectively.}
	{
		\resizebox{\textwidth}{!}{
			\begin{tabular}{c|ccccc|ccccc}
				\toprule
				\textbf{Method} & \textbf{Accuracy} & \textbf{Specificity} & \textbf{Sensitivity} & \textbf{F1} & \textbf{AUC} & \textbf{Accuracy} & \textbf{Specificity} & \textbf{Sensitivity} & \textbf{F1} & \textbf{AUC} \\
				\midrule
				&\multicolumn{5}{c|}{\textit{Fine-grained Classification}} & \multicolumn{5}{c}{\textit{Coarse-grained Classification}} \\
				\midrule
				Max-pooling& 
				46.30$_{8.07}$ & 
				89.85$_{5.66}$ &
				29.05$_{6.58}$ &
				25.72$_{7.92}$ &
				72.02$_{5.56}$ &
				56.42$_{4.77}$ &
				75.83$_{5.46}$ &
				53.73$_{6.90}$ &
				52.43$_{5.34}$ &
				79.91$_{4.68}$ \\
				Mean-pooling& 
				42.78$_{4.86}$ &
				89.48$_{5.54}$ &
				30.90$_{6.41}$& 
				26.33$_{6.54}$ &
				73.49$_{4.14}$ &
				57.21$_{5.12}$ &
				77.32$_{4.71}$ &
				53.72$_{9.18}$ &
				52.27$_{5.29}$ &
				80.15$_{5.15}$ \\
				ABMIL~\cite{ilse2018attention} & 
				51.48$_{6.66}$ &
				90.52$_{5.31}$&
				34.36$_{7.32}$& 
				30.73$_{9.56}$ &
				79.92$_{2.30}$ &
				59.26$_{5.52}$ &
				79.68$_{4.99}$ &
				58.34$_{6.37}$ &
				\textcolor{blue}{57.54$_{5.77}$} &
				85.03$_{4.25}$ \\
				CLAM-SB~\cite{lu2021data} & 
				51.35$_{6.72}$ & 
				90.26$_{6.20}$&
				37.14$_{7.10}$&
				\textcolor{blue}{34.91$_{8.47}$} &
				79.33$_{3.06}$ &
				60.28$_{4.85}$ &
				80.91$_{5.52}$ &
				\textcolor{blue}{58.68$_{6.45}$} &
				56.71$_{6.78}$ &
				85.20$_{4.85}$ \\
				CLAM-MB~\cite{lu2021data} & 
				49.70$_{6.41}$ & 
				89.46$_{5.92}$&
				36.24$_{6.36}$ &
				33.71$_{7.48}$ &
				78.30$_{3.75}$ &
				61.13$_{5.40}$ &
				81.09$_{4.72}$ &
				58.50$_{6.96}$ &
				56.53$_{5.10}$ &
				85.44$_{5.83}$ \\
				DSMIL~\cite{li2021dual} & 
				51.34$_{6.23}$ &
				90.28$_{6.61}$&
				37.31$_{6.67}$ & 
				34.90$_{5.92}$ &
				79.36$_{3.36}$ &
				\textcolor{blue}{61.79$_{4.88}$} &
				81.60$_{4.34}$ &
				58.01$_{7.47}$ &
				56.16$_{5.68}$ &
				\textcolor{blue}{85.94$_{5.45}$} \\
				TransMIL~\cite{shao2021transmil} & 
				49.28$_{6.55}$ &
				89.10$_{6.17}$&
				35.74$_{6.32}$ & 
				33.00$_{7.63}$ &
				78.23$_{3.78}$ & 
				60.58$_{5.99}$ &
				81.38$_{5.95}$ &
				56.42$_{7.30}$ &
				56.19$_{5.30}$ &
				85.20$_{5.52}$ \\
				HIPT~\cite{chen2022scaling} & 
				49.50$_{6.28}$ & 
				89.71$_{6.04}$ &
				35.17$_{7.28}$ &
				34.87$_{8.22}$ &
				79.34$_{3.26}$ & 
				61.45$_{5.03}$ &
				81.28$_{5.66}$ &
				56.20$_{6.95}$ &
				56.02$_{5.24}$ &
				85.05$_{5.30}$ \\
				DTFD~\cite{zhang2022dtfd} & 
				49.38$_{6.43}$ &
				88.03$_{6.10}$ &
				36.61$_{6.41}$ &  
				32.70$_{7.86}$ &
				78.41$_{3.71}$ &
				61.28$_{5.18}$ &
				81.14$_{4.73}$ &
				57.22$_{6.97}$ &
				56.96$_{5.17}$ &
				84.96$_{5.81}$ \\
				PMIL~\cite{yu2023prototypical} & 
				48.89$_{6.44}$ &
				88.20$_{7.12}$&
				35.02$_{6.15}$ & 
				34.21$_{3.53}$ &
				77.53$_{3.33}$ &
				61.37$_{5.13}$ &
				81.27$_{5.66}$ &
				56.21$_{6.96}$ &
				56.03$_{6.25}$ &
				84.66$_{5.31}$ \\
				PseMix~\cite{liu10385148} & 
				51.41$_{5.94}$ &
				90.13$_{6.02}$&
				35.84$_{6.92}$ & 
				33.59$_{5.68}$ &
				78.70$_{5.05}$ &
				61.59$_{6.16}$&
				\textcolor{blue}{82.14$_{3.14}$}&
				57.60$_{6.70}$&
				56.37$_{5.39}$& 
				85.32$_{4.05}$\\
				ACMIL~\cite{zhang2023attentionchallenging} & 
				46.52$_{4.51}$ &
				87.16$_{5.02}$&
				35.38$_{7.03}$ & 
				32.26$_{6.44}$ &
				77.85$_{3.06}$ &
				61.11$_{5.87}$&
				82.06$_{5.21}$&
				57.82$_{5.10}$&
				57.22$_{5.16}$& 
				85.60$_{4.39}$\\
				S4MIL~\cite{fillioux2023structured} & 
				\textcolor{blue}{52.40$_{6.78}$}& 
				\textcolor{blue}{90.60$_{5.94}$}&
				\textcolor{blue}{37.98$_{7.94}$} &
				34.80$_{9.71}$ &
				\textcolor{blue}{80.20$_{3.10}$} & 61.31$_{4.88}$ &
				81.41$_{4.59}$ &
				58.18$_{6.65}$ &
				57.08$_{5.30}$ &
				85.34$_{4.78}$ \\
				HMIL (Ours) & 
				\textcolor{red}{55.56$_{5.92}$} & 
				\textcolor{red}{91.21$_{5.90}$}&
				\textcolor{red}{38.61$_{6.02}$} &
				\textcolor{red}{38.98$_{7.21}$} &
				\textcolor{red}{83.03$_{2.85}$} &
				\textcolor{red}{64.07$_{5.12}$} & 
				\textcolor{red}{84.49$_{4.28}$}&
				\textcolor{red}{60.79$_{5.13}$} &
				\textcolor{red}{58.75$_{5.21}$} &
				\textcolor{red}{87.66$_{4.52}$} \\
				\bottomrule
	\end{tabular}}}
	\label{tab:results_1}
\end{table*}

\begin{table*}[h]
	\centering
	\caption{Evaluation of performance on the histology WSI datasets PANDA. We report the results in the form of $\text{Mean}_{\text{Std}}$. The best and second best results are highlighted in \textcolor{red}{red} and \textcolor{blue}{blue}, respectively.}
	{
		\resizebox{\textwidth}{!}{
			\begin{tabular}{c|ccccc|ccccc}
				\toprule
				\textbf{Method} & \textbf{Accuracy} & \textbf{Specificity} & \textbf{Sensitivity} & \textbf{F1} & \textbf{AUC} & \textbf{Accuracy} & \textbf{Specificity} & \textbf{Sensitivity} & \textbf{F1} & \textbf{AUC} \\
				\midrule
				&\multicolumn{5}{c|}{\textit{Fine-grained Classification}} & \multicolumn{5}{c}{\textit{Coarse-grained Classification}} \\
				\midrule
				Max-pooling& 
				61.21$_{1.50}$ &
				90.74$_{1.75}$ &
				54.47$_{1.82}$ & 
				54.47$_{2.01}$ &
				88.22$_{0.68}$ & 
				76.08$_{1.48}$ &
				86.78$_{0.80}$ &
				70.19$_{1.95}$ &
				70.98$_{2.06}$ &
				88.99$_{0.87}$\\
				Mean-pooling& 
				61.51$_{1.49}$ &
				90.65$_{1.84}$ &
				55.21$_{1.64}$ & 
				55.43$_{1.67}$ &
				88.32$_{0.71}$ & 
				76.48$_{1.43}$ &
				87.03$_{0.74}$ &
				70.97$_{1.61}$ &
				72.03$_{1.73}$ &
				89.48$_{0.85}$ \\
				ABMIL~\cite{ilse2018attention} & 
				62.06$_{1.59}$ & 
				90.59$_{2.43}$ &
				\textcolor{blue}{55.75$_{2.01}$} &
				\textcolor{blue}{55.87$_{2.08}$} &
				88.34$_{0.60}$ & 
				76.28$_{1.45}$ &
				86.97$_{0.93}$ &
				70.60$_{1.50}$ &
				71.48$_{1.63}$ &
				89.24$_{0.51}$\\
				CLAM-SB~\cite{lu2021data} & 
				61.25$_{1.76}$ & 
				90.56$_{1.92}$ &
				54.67$_{2.14}$&
				54.49$_{2.49}$ &
				88.14$_{0.48}$ & 
				76.51$_{1.32}$ &
				87.00$_{0.73}$ &
				70.81$_{1.62}$ &
				71.63$_{1.71}$ &
				89.20$_{0.62}$\\
				CLAM-MB~\cite{lu2021data} & 
				61.70$_{1.76}$ &  
				90.61$_{1.87}$ &
				55.24$_{1.58}$ &
				55.20$_{2.43}$ &
				88.26$_{0.62}$ & 
				\textcolor{blue}{76.73$_{1.53}$} &
				87.03$_{0.85}$ &
				70.87$_{2.02}$ &
				\textcolor{blue}{72.07$_{2.08}$} &
				\textcolor{blue}{89.51$_{0.74}$}\\
				DSMIL~\cite{li2021dual} & 
				61.74$_{1.46}$ &
				\textcolor{blue}{91.22$_{1.51}$} &
				55.25$_{1.62}$ & 
				55.42$_{1.68}$ &
				\textcolor{blue}{88.44$_{0.68}$} & 76.52$_{1.32}$ &
				86.93$_{0.71}$ &
				70.85$_{1.58}$ &
				71.67$_{1.74}$ &
				89.32$_{0.77}$ \\
				TransMIL~\cite{shao2021transmil} & 
				61.40$_{1.55}$ &
				90.87$_{1.56}$&
				54.78$_{1.90}$ & 
				54.70$_{2.14}$ &
				88.23$_{0.72}$ & 
				76.25$_{1.51}$ &
				86.98$_{0.83}$ &
				70.62$_{1.96}$ &
				71.36$_{2.00}$ &
				88.98$_{0.80}$\\
				HIPT~\cite{chen2022scaling} & 
				61.28$_{1.53}$ & 
				90.79$_{2.13}$ &
				54.20$_{2.25}$ &
				54.17$_{2.49}$ &
				88.28$_{0.45}$ & 
				76.43$_{1.40}$ &
				86.92$_{0.76}$ &
				70.61$_{1.31}$ &
				71.34$_{1.51}$ &
				89.14$_{0.88}$ \\
				DTFD~\cite{zhang2022dtfd} & 
				61.56$_{1.57}$ &
				91.02$_{2.06}$ &
				54.93$_{1.88}$ & 
				54.84$_{2.10}$ &
				88.29$_{0.72}$ &
				76.36$_{1.52}$ &
				87.05$_{0.83}$ &
				70.77$_{1.93}$ &
				71.49$_{1.98}$ &
				89.07$_{0.94}$\\
				PMIL~\cite{yu2023prototypical} & 
				61.26$_{1.67}$ &
				90.14$_{2.47}$&
				54.66$_{2.17}$ & 
				54.56$_{2.22}$ &
				88.05$_{0.57}$ & 
				76.32$_{1.50}$ &
				86.79$_{0.85}$ &
				70.65$_{1.34}$ &
				71.45$_{2.15}$ &
				88.93$_{0.94}$\\
				PseMix~\cite{liu10385148} & 
				\textcolor{blue}{62.14$_{1.73}$}&
				90.80$_{1.30}$&
				55.71$_{1.30}$&
				55.42$_{1.91}$&
				88.37$_{0.95}$&
				76.61$_{1.67}$&
				87.07$_{0.92}$&
				\textcolor{blue}{70.93$_{2.43}$}&
				71.92$_{2.24}$&
				89.47$_{0.48}$\\
				ACMIL~\cite{zhang2023attentionchallenging} & 
				61.56$_{1.89}$ &
				90.45$_{1.36}$&
				54.61$_{2.20}$ & 
				54.28$_{2.63}$ &
				88.40$_{0.72}$ &
				76.68$_{1.60}$&
				\textcolor{blue}{87.89$_{0.72}$}&
				70.45$_{1.51}$&
				71.17$_{1.57}$& 
				89.43$_{0.92}$\\
				S4MIL~\cite{fillioux2023structured} & 
				61.47$_{1.57}$ &  
				91.14$_{1.41}$&
				54.31$_{1.90}$ &
				54.94$_{2.06}$ &
				88.30$_{0.71}$ & 
				76.30$_{1.50}$ &
				86.71$_{0.74}$ &
				70.64$_{1.44}$ &
				71.39$_{2.06}$ &
				89.05$_{0.62}$\\
				HMIL (Ours) & 
				\textcolor{red}{63.41$_{1.42}$} & 
				\textcolor{red}{92.42$_{1.54}$} &
				\textcolor{red}{58.36$_{1.57}$} &
				\textcolor{red}{58.16$_{1.70}$} &
				\textcolor{red}{89.43$_{0.27}$} &
				\textcolor{red}{77.23$_{1.38}$} & 
				\textcolor{red}{88.02$_{0.77}$}&
				\textcolor{red}{73.08$_{1.12}$} &
				\textcolor{red}{73.50$_{1.37}$} &
				\textcolor{red}{90.25$_{0.84}$} \\
				\bottomrule
	\end{tabular}}}
	\label{tab:results_2}
\end{table*}

\begin{table*}[ht] 
	\caption{Performance evaluation on the cytology WSI dataset CCC. We report the results in the form of $\text{Mean}_{\text{Std}}$. The best and second best results are highlighted in \textcolor{red}{red} and \textcolor{blue}{blue}, respectively.}
	\centering
	\resizebox{\textwidth}{!}{
		\begin{tabular}{c|ccccc|ccccc}
			\toprule
			{\textbf{Method}} & {\textbf{Accuracy}} & \textbf{Specificity} & \textbf{Sensitivity} & {\textbf{F1}} & {\textbf{AUC}} & {\textbf{Accuracy}} & \textbf{Specificity} & \textbf{Sensitivity} & {\textbf{F1}} & {\textbf{AUC}}\\
			\midrule
			&\multicolumn{5}{c|}{\textit{Fine-grained Classification}} & \multicolumn{5}{c}{\textit{Coarse-grained Classification}} \\
			\midrule
			Max-pooling& 
			72.57$_{0.25}$ & 
			87.93$_{0.21}$ & 
			25.08$_{1.42}$ & 
			25.62$_{0.57}$ &
			81.29$_{0.42}$ &
			81.69$_{0.64}$ & 
			84.26$_{0.39}$ & 
			81.53$_{0.75}$ & 
			81.59$_{0.77}$ &
			87.88$_{0.46}$ \\
			Mean-pooling& 
			73.13$_{0.32}$ & 
			85.21$_{0.97}$ & 
			25.76$_{0.96}$ & 
			25.87$_{0.35}$ &
			82.75$_{0.45}$ &
			84.17$_{0.32}$ & 
			83.72$_{0.97}$ & 
			82.97$_{0.96}$ & 
			82.45$_{0.35}$ &
			88.12$_{0.45}$\\
			ABMIL~\cite{ilse2018attention} & 
			76.61$_{0.47}$ & 
			88.85$_{0.21}$ & 
			26.93$_{0.17}$ & 
			26.03$_{0.96}$ &
			87.26$_{0.16}$ &
			83.77$_{0.59}$ & 
			83.56$_{0.67}$ & 
			83.79$_{0.98}$ & 
			83.74$_{1.09}$ &
			89.38$_{0.61}$\\
			CLAM-SB~\cite{lu2021data} & 
			74.16$_{0.58}$ & 
			86.73$_{0.10}$ & 
			36.06$_{0.93}$ & 
			34.68$_{0.22}$ &
			85.80$_{0.55}$ &
			81.69$_{0.92}$ & 
			84.86$_{0.61}$ & 
			82.80$_{1.40}$ & 
			82.68$_{0.62}$ &
			88.32$_{0.58}$ \\
			CLAM-MB~\cite{lu2021data} & 
			74.81$_{0.94}$ & 
			87.52$_{0.21}$ & 
			\textcolor{blue}{37.92$_{0.18}$} & 
			36.29$_{0.95}$ &
			86.09$_{0.12}$ & 
			85.30$_{0.99}$ & 
			88.19$_{0.36}$ & 
			85.12$_{0.80}$ & 
			85.21$_{0.97}$ &
			90.80$_{0.37}$ \\
			DSMIL~\cite{li2021dual} & 
			75.18$_{0.76}$ & 
			88.29$_{0.88}$ & 
			29.77$_{0.52}$ & 
			28.44$_{0.72}$ &
			87.29$_{0.80}$ & 
			86.41$_{0.44}$ & 
			88.19$_{0.69}$ & 
			86.29$_{1.12}$ & 
			86.34$_{0.80}$ &
			91.57$_{0.94}$ \\
			TransMIL~\cite{shao2021transmil} & 
			74.96$_{0.92}$ & 
			86.96$_{0.35}$ & 
			28.39$_{0.66}$ & 
			32.27$_{0.54}$ &
			84.05$_{0.39}$ & 
			81.59$_{0.30}$ & 
			83.86$_{0.35}$ & 
			81.80$_{1.22}$ & 
			81.68$_{1.15}$ &
			86.54$_{0.35}$ \\
			HIPT~\cite{chen2022scaling} & 
			77.11$_{0.62}$ & 
			91.38$_{0.14}$ & 
			35.23$_{0.43}$ & 
			39.24$_{1.57}$ &
			87.03$_{0.48}$ & 
			84.87$_{0.70}$ & 
			88.72$_{0.94}$ & 
			84.63$_{0.62}$ & 
			84.75$_{0.67}$ &
			91.34$_{1.06}$ \\
			DTFD~\cite{zhang2022dtfd} & 
			74.51$_{0.49}$ & 
			85.94$_{0.37}$ & 
			35.59$_{0.81}$ & 
			34.14$_{0.87}$ &
			85.96$_{0.41}$ & 
			86.98$_{0.96}$ & 
			88.54$_{0.63}$ & 
			85.66$_{0.69}$ & 
			85.82$_{0.72}$ &
			92.10$_{0.36}$ \\
			PMIL~\cite{yu2023prototypical} & 
			76.32$_{0.49}$ & 
			88.19$_{0.31}$ & 
			34.24$_{0.83}$ & 
			35.26$_{0.28}$ &
			84.07$_{0.16}$ & 
			81.78$_{0.49}$ & 
			83.92$_{0.80}$ & 
			82.58$_{0.36}$ & 
			82.62$_{0.31}$ &
			88.09$_{0.62}$ \\
			PseMix~\cite{liu10385148} & 
			76.07$_{0.79}$&
			87.24$_{0.26}$&
			33.32$_{0.61}$&
			32.24$_{0.39}$&
			87.31$_{0.24}$&
			\textcolor{blue}{88.09$_{0.56}$}&
			88.78$_{0.84}$&
			85.47$_{0.72}$&
			82.79$_{0.62}$&
			92.02$_{0.38}$\\
			ACMIL~\cite{zhang2023attentionchallenging} & 
			76.56$_{1.89}$ &
			90.42$_{1.36}$&
			33.61$_{2.20}$ & 
			38.28$_{2.63}$ &
			87.47$_{0.72}$ &
			86.68$_{1.60}$&
			88.89$_{0.72}$&
			85.45$_{1.51}$&
			86.17$_{1.57}$& 
			91.35$_{0.92}$\\
			S4MIL~\cite{fillioux2023structured} & 
			\textcolor{blue}{77.30$_{0.88}$} & 
			\textcolor{blue}{92.54$_{0.11}$} & 
			35.09$_{0.41}$ & 
			\textcolor{blue}{40.87$_{0.58}$} &
			\textcolor{blue}{87.64$_{0.43}$} & 
			87.39$_{0.22}$ & 
			\textcolor{blue}{88.93$_{0.53}$} & 
			\textcolor{blue}{87.29$_{0.39}$} & 
			\textcolor{blue}{87.33$_{0.35}$} &
			\textcolor{blue}{92.70$_{0.16}$} \\
			HMIL (Ours) & 
			\textcolor{red}{80.25$_{0.32}$} & 
			\textcolor{red}{93.93$_{0.12}$} & 
			\textcolor{red}{40.97$_{0.92}$} & 
			\textcolor{red}{44.39$_{0.99}$} &
			\textcolor{red}{91.24$_{0.18}$} &
			\textcolor{red}{91.44$_{0.23}$} & 
			\textcolor{red}{89.32$_{0.27}$} & 
			\textcolor{red}{89.39$_{0.81}$} & 
			\textcolor{red}{89.66$_{0.90}$} &
			\textcolor{red}{95.88$_{0.17}$} \\
			\bottomrule
	\end{tabular}}
	
	\hfill
	\label{tab:results_3}
\end{table*}

\subsection{Experiment Results and Ablation Studies}
\subsubsection{Fine-grained Classification}
We evaluated the proposed HMIL in fine-grained WSI classification tasks and summarized the results in the left part of Tables \ref{tab:results_1}-\ref{tab:results_3}. From the results, it is observed our proposed HMIL outperforms all compared methods in terms of accuracy, specificity, sensitivity, F1 score, and AUC, demonstrating its effectiveness in identifying subtle differences and patterns within WSI images. Specifically, in the histology BRACS dataset (Table \ref{tab:results_1}), HMIL achieved the highest accuracy of $55.56 \pm 5.92\%$, specificity of $91.21 \pm 5.90\%$, sensitivity of $38.61 \pm 6.02\%$, F1 score of $38.98 \pm 7.21\%$, and AUC of $83.03 \pm 2.85$. Similarly, for the histology PANDA dataset (Table \ref{tab:results_2}), HMIL demonstrated superior performance with an accuracy of $63.41 \pm 1.42\%$, specificity of $92.42 \pm 1.54\%$, sensitivity of $58.36 \pm 1.57\%$, F1 score of $58.16 \pm 1.70\%$, and AUC of $89.43 \pm 0.27\%$. Lastly, in the CCC dataset (Table \ref{tab:results_3}), which is more challenging inferred from the metrics, HMIL outperformed other methods with an accuracy of $80.25 \pm 0.32\%$, specificity of $93.93 \pm 0.12\%$, sensitivity of $40.97 \pm 0.92\%$, F1 score of $44.39 \pm 0.99\%$, and AUC of $91.24 \pm 0.18\%$.
\begin{figure*}[h] 
	\centering
	\includegraphics[width=\textwidth]{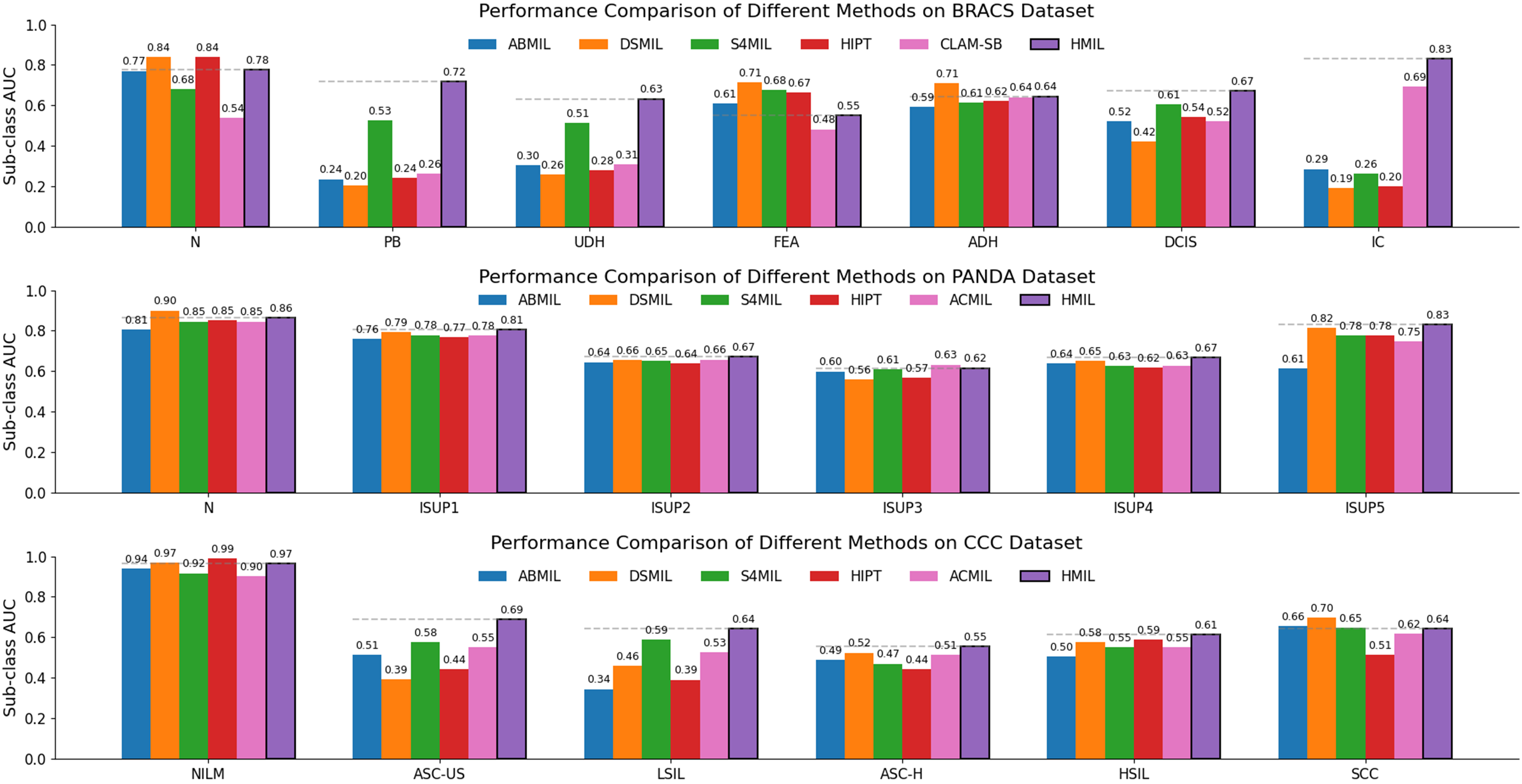}
	\caption{The class-wise AUC distribution of top-performing methods on BRACS (Top), PANDA (Middle), and CCC (Bottom) datasets.}\label{fig:classwise}
\end{figure*}

\begin{figure*}[h] 
	\centering
	\includegraphics[width=\textwidth, height=0.6\textheight]{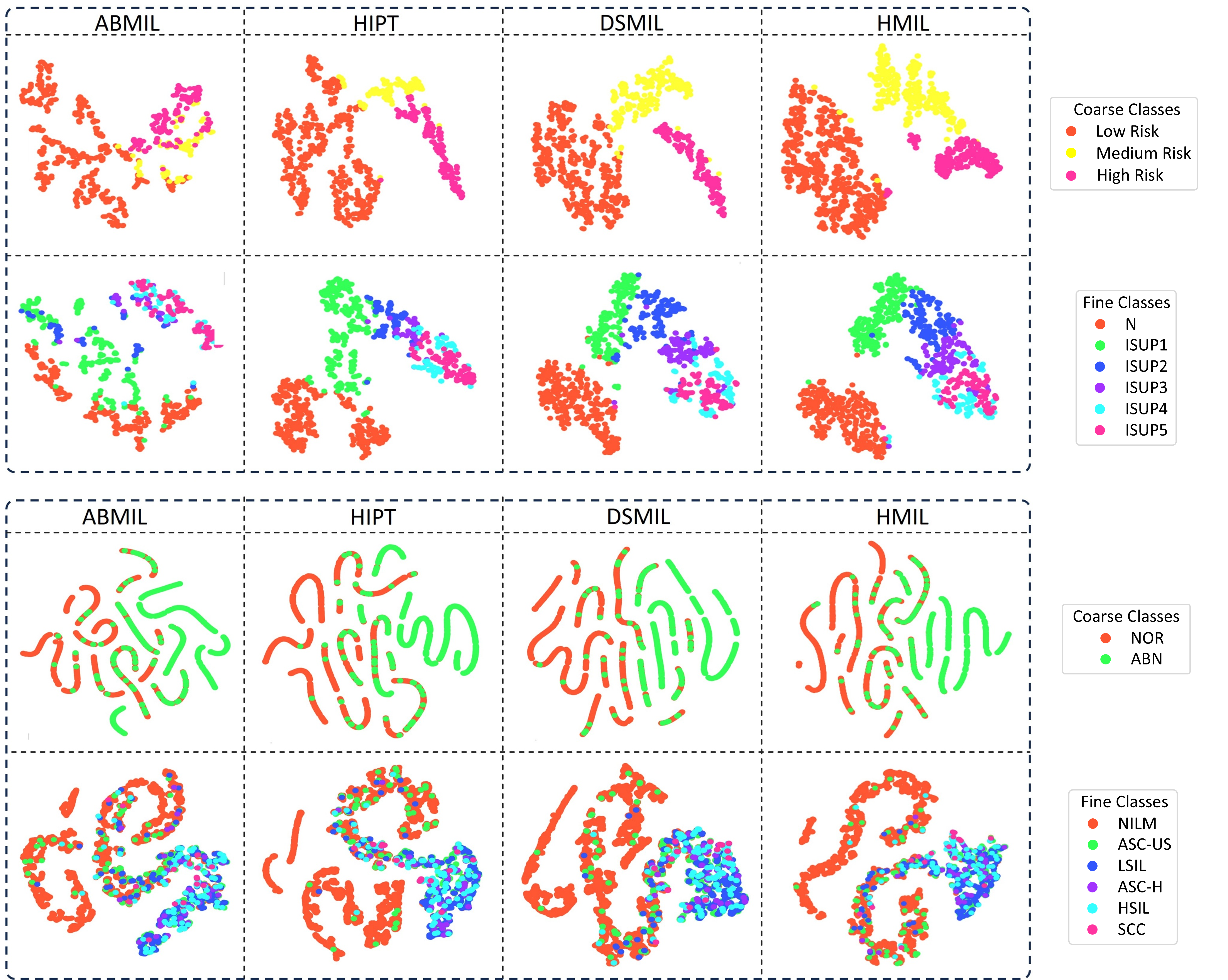}
	\caption{The t-SNE visualization on PANDA (top) and CCC (bottom) datasets. The upper section of each dataset displays coarse-grained classes, while the lower section showcases fine-grained classes. }\label{fig:tsne}
\end{figure*}

The quantitative results demonstrate that current MIL methods still face challenges in fine-grained classification tasks, as indicated by the relatively low sensitivity metric. Methods that rely solely on learning attention from each instance may not be sufficient to discern subtle differences. 

While the ABMIL method shows stability through attention-based classification, more complex designs, such as CLAM-SB and CLAM-MB, which utilize learned multi-branch class-wise clusters, achieve no significant improvement compared to their single-branch variant, CLAM-SB, and even exhibit worse performance on the BRACS dataset. Data augmentation approaches like DTFD and PseMix have improved sensitivity, but at the cost of reduced specificity, and they have not demonstrated significant advantages in enhancing overall model performance.  Both DSMIL and HIPT benefited from their pretrained encoders. However, HIPT's pretrained weights are based on TCGA datasets \cite{chen2022scaling}, introducing significant domain shift issues, while DSMIL has a relatively small pretraining size and less effective aggregator, as highlighted in the ablation studies in Sect. \ref{sec: ablation}. S4MIL, which leverages state space model architecture, achieves nearly the second-best performance but still falls short of the proposed HMIL. This further underscores the advantage of the supervision provided by label hierarchy in fine-grained WSI classification tasks.

\subsubsection{Coarse-grained Classification} We also explored whether hierarchical alignment leads to mutual enhancement by conducting coarse-grained classification experiments. HMIL exhibits significant improvements compared to baseline methods. In the BRACS dataset (Table \ref{tab:results_1}), HMIL achieved an accuracy of $64.07 \pm 5.12$\%, specificity of $84.49 \pm 4.28$\%, sensitivity of $60.79 \pm 5.13$\%, F1 score of $58.75 \pm 5.21$\%, and AUC of $87.66 \pm 4.52$\%. In the PANDA dataset (Table \ref{tab:results_2}), HMIL attained the highest accuracy of $77.23 \pm 1.38$\%, specificity of $88.02 \pm 0.77$\%, sensitivity of $73.08 \pm 1.12$\%, F1 score of $73.50 \pm 1.37$\%, and AUC of $90.25 \pm 0.84$. Finally, for the CCC dataset (Table \ref{tab:results_3}), HMIL achieved an accuracy of $91.44 \pm 0.23$\%, specificity of $89.32 \pm 0.27$\%, sensitivity of $89.39 \pm 0.81$\%, F1 score of $89.66 \pm 0.90$, and AUC of $95.88 \pm 0.17$.

These results confirm the utility of label hierarchy in facilitating classification tasks at the coarse level. In coarse-grained tasks, where features are more distinguishable, the fine branch through hierarchical alignment serves to confirm and refine feature representation, thereby enhancing overall accuracy. Since the classification task is easier with fewer categories to identify, methods with learned attention from the instances like CLAM-MB, with its multi-branch class-wise clusters, show improved performance compared to their single-branch variant, CLAM-SB. Other methods also show varying degrees of improvement. However, they still fall short of our HMIL. This consistent performance across different datasets underscores the versatility and effectiveness of HMIL.

\subsubsection{Class-wise Performance Visualization} We present the class-wise AUC distribution and bag feature visualization for the top-performing methods in Figures \ref{fig:classwise} and \ref{fig:tsne}. From the class-wise AUC distributions, a notable observation is that although pretraining methods like DSMIL and HIPT exhibit high overall performance, they tend to perform better on classes with larger sample sizes. In contrast, the other baselines yield more balanced results, particularly S4MIL. Nevertheless, our method not only achieves a more balanced performance but also demonstrates superior overall results, which we attribute to the contextual guidance provided by hierarchical context.

To observe and visualize the effectiveness of feature representation, we employ the t-SNE method \cite{van2008visualizing} to visualize the learned bag features $\mathbf{B_{\{c,f\}}}$ at each branch of HMIL. Additionally, we compare the feature representation capabilities of ABMIL, DSMIL, and HIPT in coarse- and fine-grained WSI classification using the PANDA and CCC datasets, as these datasets provide a sufficient sample size for effective visualization. In the results for the PANDA dataset, the upper section illustrates that ABMIL exhibits minimal clustering and lacks distinct separation among coarse-grained categories. In contrast, HIPT and DSMIL, having benefited from pretraining, show improved feature representation; however, some degree of overlap persists in their clustering. Notably, HMIL leverages contextual guidance to achieve a significantly clearer separation among coarse-grained categories, underscoring its effectiveness in distinguishing between different risk levels. When we examine fine-grained classifications, the challenges become more pronounced. DSMIL and HIPT exhibit significant overlap in fine-grained tasks, highlighting the challenges of classification. In contrast, HMIL demonstrates a better ability to distinguish between different ISUP categories. Similar observations are noted within the cytology CCC dataset, which presents even greater challenges for classification, reinforcing the consistency of our findings. Collectively, these results underscore the superior feature representation capabilities of HMIL in both coarse and fine-grained classifications, particularly in addressing the complexities inherent in fine-grained tasks. This positions HMIL as a particularly effective model for managing intricate datasets.

\subsubsection{Ablation Studies}\label{sec: ablation}
\begin{figure}[h] 
	\centering
	\includegraphics[width=0.45\textwidth]{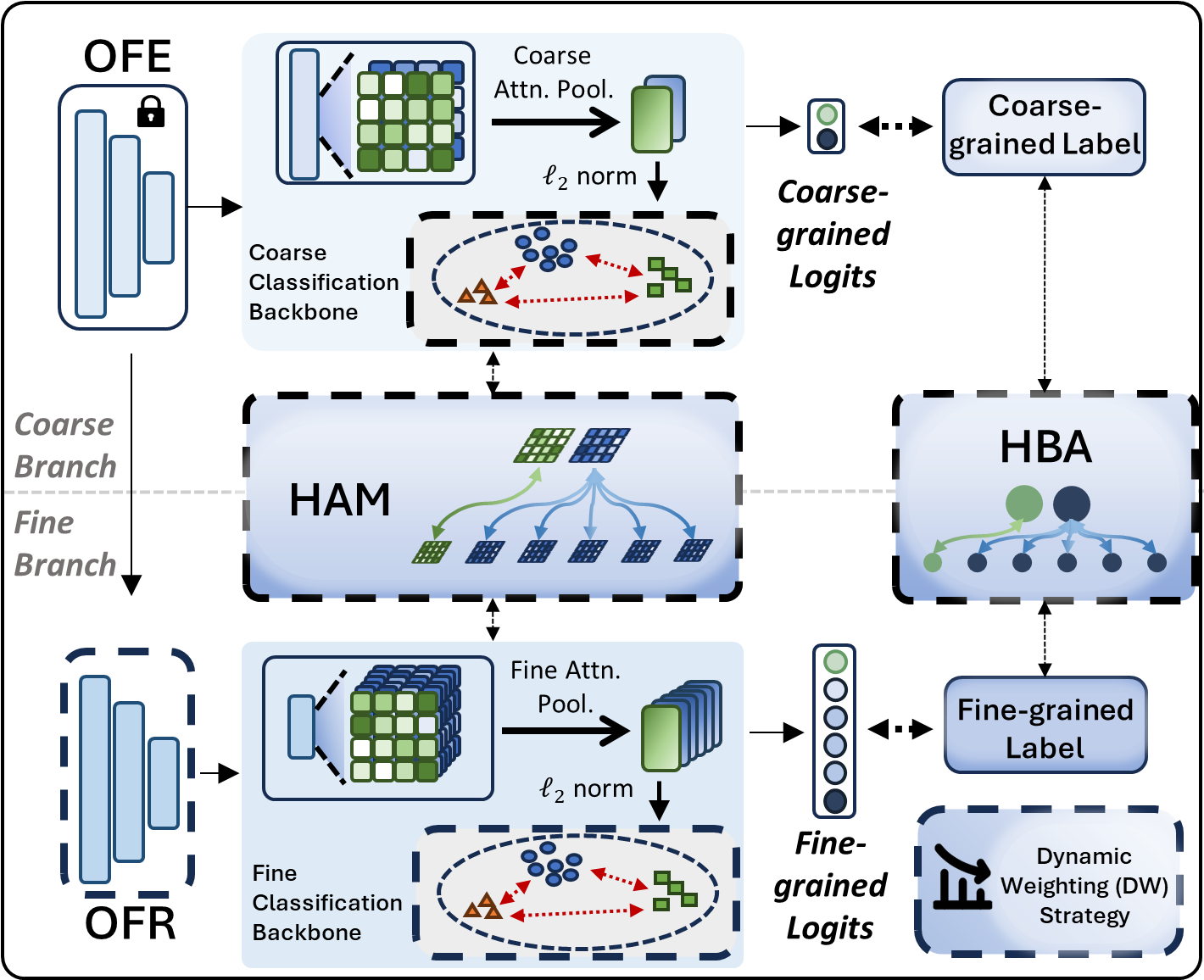}
	\caption{Ablation study conducted on the HMIL framework. The modules and strategy involved in the study, namely HAM, HBA, OFR, SCL in different branches, and DW, are delineated with dashed lines.}\label{fig:ablation}
\end{figure}

To further study the efficacy of our HMIL architecture, as illustrated in Figure \ref{fig:ablation}, we conduct a comprehensive analysis using the test set of the three evaluated datasets and report the results in terms of AUC for one-versus-rest classification scenarios.

\textbf{Holistic Hierarchical Guidance Matters in Fine-grained WSI Classification.} We first study the effectiveness of hierarchical guidance at different MIL levels within our HMIL framework, focusing on instance-level guidance via hierarchical attention mapping (HAM), bag-level guidance via hierarchical bag alignment (HBA), and their combination. From Table \ref{ablation_1}, we note that starting with only the fine branch using a class-wise attention learning mechanism, similar to ABMIL but with added class-wise constraints, leads to degraded performance. Without any guidance provided by the hierarchical mapping, the performance become worser when a coarse branch is added. While hierarchical instance-level guidance offers moderate improvements, it remains inferior to the flat fine branch model. In contrast, combining the coarse branch with bag-level guidance surpasses the flat fine branch. The best performance is achieved by integrating both instance-level and bag-level guidance with the coarse branch, highlighting their complementary strengths. These results underscore the importance of combining alignment strategies to capture hierarchical relationships and enhance classification accuracy.

\begin{table}[h] 
	\centering
	\caption{Evaluation of hierarchical guidance on model performance. \cmark denotes applying the corresponding module to the model. Best results are highlighted in bold.}
	\resizebox{0.49\textwidth}{!}{
		\begin{tabular}{cccc|ccc}
			\toprule
			\textbf{FB} & \textbf{CB} & \textbf{HAM} & \textbf{HBA} & \textbf{BRACS} & \textbf{PANDA} & \textbf{CCC} \\
			\midrule
			\cmark &    &     &     & 78.03 & 88.01 & 87.48 \\
			\cmark & \cmark &     &     & 76.83 & 87.62 & 87.43 \\
			\cmark & \cmark & \cmark &     & 77.92 & 87.45 & 86.64 \\
			\cmark & \cmark &  & \cmark    &  79.23 & 88.55 & 87.06 \\
			\cmark & \cmark & \cmark & \cmark & \textbf{81.22} & \textbf{89.03} & \textbf{89.52} \\
			\midrule
			\multicolumn{4}{c|}{ABMIL\cite{ilse2018attention}} & 79.84 & 88.32 & 88.59 \\
			\bottomrule
	\end{tabular}}
	\label{ablation_1}
\end{table}

We next examine the contribution of our hierarchical feature refinement (HFE) components, including the online feature re-embedding (OFR) module, supervised contrastive learning (SCL) at different branches, and the dynamic weighting (DW) strategy upon the core model, which operates in dual-branch with holistic hierarchical alignment, the results are summarized in Table \ref{ablation_2}.

\begin{table}[h]
	\centering
	\caption{Comparison of our approach using different combinations of the proposed modules OFR, DW, and SCL. \cmark denotes applying the corresponding module to the model. Subscripts $ f $ and $ c $ denote applying the SCL module to the fine or coarse branch, respectively. Best results are highlighted in bold.}
	\resizebox{0.49\textwidth}{!}{
		\begin{tabular}{ccccc|ccc}
			\toprule
			\textbf{Core} & \textbf{OFR} & \textbf{SCL$_f$} & \textbf{DW} &  \textbf{SCL$_c$} & \textbf{BRACS} & \textbf{PANDA} & \textbf{CCC} \\
			\midrule
			\cmark & & & & & 80.47 & 88.65 & 88.84 \\
			\midrule
			\cmark & \cmark & & & & 81.22 & 89.03 & 89.52 \\
			\cmark & & \cmark & & & 80.94 & 88.92 & 89.02 \\
			\cmark &  & & \cmark & & 81.06 & 89.07 & 89.21 \\
			\midrule
			\cmark & \cmark & \cmark & & & 81.39 & 89.18 & 89.94 \\
			\cmark & \cmark & & \cmark & & 82.19 & 89.33 & 90.95 \\
			\cmark & & \cmark & \cmark & & 81.83 & 89.25 & 90.43 \\
			\midrule
			\cmark & \cmark & \cmark & \cmark & & \textbf{83.42} & \textbf{89.39} & \textbf{91.16} \\
			\cmark & \cmark & \cmark & \cmark & \cmark & 83.36 & 89.35 & 91.14 \\
			\bottomrule
	\end{tabular}} \label{ablation_2}
\end{table}

From the results, we observe that the hierarchical feature refinement components each contribute to enhancing the model's performance. The OFR module improves feature representations for the fine branch, while the DW strategy balances information from both branches. The SCL module also provides performance gains. When combined, these components work synergistically, with the highest performance achieved when all three are used together, demonstrating their collective effectiveness in refining features and balancing information for superior classification performance. For a comprehensive evaluation, we also applied SCL to the coarse branch. Notably, the results did not show significant improvement, suggesting that SCL is more effective in fine-grained contexts. This further reinforces our understanding that the primary benefits of SCL are realized when applied to the fine branch.

Finally, we conducted ablation studies on loss function as shown in Table \ref{ablation_5} to verify the effectiveness of our dynamic weighting strategy based on the following loss function:

\begin{equation}
	\begin{aligned}
		\mathcal{L} &= a \cdot (\mathcal{L}^{(f)}_{ce} + \mathcal{L}_{ia} + \mathcal{L}_{ba} ) + b \cdot \mathcal{L}_{reg} + \mathcal{L}^{(c)}_{ce} 
	\end{aligned}
\end{equation}

\begin{table}[h]
	\centering
	\caption{Ablation studies of the proposed HMIL framework for loss function, with the best results highlighted in bold.}
	\resizebox{0.4\textwidth}{!}{
		\begin{tabular}{ccc|ccc}
			\toprule
			\textbf{$a$} & \textbf{$b$} & \textbf{$\tau$} & \textbf{BRACS} & \textbf{PANDA} & \textbf{CCC}  \\
			\midrule
			1 & 1 & 0.1 & 82.66 & 89.16 & 91.05 \\
			1 & 0.1 & 0.1 & 81.85 & 88.51 & 89.72 \\
			1 & 0.01 & 0.1 & 81.46  & 88.27 & 87.88 \\ 
			1 & 0 & 0.1 & 80.95 & 88.14 & 87.82 \\
			0.1 & 1 & 0.1 & 80.81 & 88.32 & 88.35 \\  
			0.01 & 1 & 0.1 & 79.92 & 88.24 & 87.69 \\
			0 & 1 & 0.1 & 79.65 & 88.06 & 87.29 \\
			\midrule
			\multicolumn{2}{c}{Dynamic Weighting} & 1 & 82.76 & 88.93 & 90.55 \\
			\multicolumn{2}{c}{Dynamic Weighting} & 0.01 & 83.46 & 89.14 & 90.53 \\
			\multicolumn{2}{c}{Dynamic Weighting} & 0.1 & \textbf{83.42} & \textbf{89.39} & \textbf{91.16} \\
			\bottomrule
	\end{tabular}}
	\label{ablation_5}
\end{table}

The results indicate that the best performance is achieved with a combination of dynamic weighting and proposed temperature parameter highlighted in bold. Notably, the dynamic weighting approach consistently outperforms static configurations, demonstrating its ability to enhance classification accuracy across all datasets. This underscores the importance of adaptive loss functions in optimizing model performance within our HMIL framework.

\textbf{Hierarchical Guidance Has Mutual Benifits.} In addition to fine-grained WSI classification, to comprehensively study the effect of hierarchical alignment for coarse-grained WSI classification, we also conduct an ablation study to explore the hierarchical guidance at different MIL levels and the effectiveness of HFE components as detailed in Table \ref{ablation_3}. It should be noted that under this setting, models with DW strategy utilize the following loss function, which concentrating on the coarse branch, for balancing the knowledge from each branch:

\begin{equation}
	\begin{aligned}
		\mathcal{L} &= \beta \cdot (\mathcal{L}^{(f)}_{ce} + \mathcal{L}_{ia} + \mathcal{L}_{ba} ) + (1 - \beta) \cdot \mathcal{L}_{reg} + \mathcal{L}^{(c)}_{ce} 
	\end{aligned}
\end{equation}

\begin{table}[h]
	\centering
	\caption{Ablation studies of the proposed HMIL framework for coarse-grained cancer subtyping task, with the best results highlighted in bold.}
	\resizebox{0.49\textwidth}{!}{
		\begin{tabular}{ccccc|ccc}
			\toprule
			\textbf{CB} & \textbf{FB} & \textbf{HAM} & \textbf{HBA} & \textbf{HFE} & \textbf{BRACS} & \textbf{PANDA} & \textbf{CCC}  \\
			\midrule
			\cmark & & & & & 85.15 & 89.17 & 90.42 \\
			\cmark & \cmark & & & & 87.34 & 90.01 & 94.92 \\
			\cmark & \cmark & \cmark & & & 87.38  & 89.94 & 95.01 \\ 
			\cmark & \cmark & &\cmark & & 87.45 & 90.05 & 95.26 \\
			\cmark & \cmark & \cmark & \cmark & & 87.48 & 90.19 & 95.60 \\  
			\cmark & \cmark & \cmark & \cmark & \cmark & \textbf{87.49} & \textbf{90.22} & \textbf{95.65} \\
			\midrule
			\multicolumn{5}{c|}{ABMIL\cite{ilse2018attention}} & 83.92 & 89.26 & 89.56 \\
			\multicolumn{5}{c|}{HIPT\cite{chen2022scaling}} & 86.27 & 89.17 & 90.51 \\
			\multicolumn{5}{c|}{S4MIL\cite{fillioux2023structured}} & 86.38 & 89.01 & 92.87 \\
			\bottomrule
	\end{tabular}}
	\label{ablation_3}
\end{table}

From the results, we observe that introducing the fine branch without applying any alignment strategies already leads to a noticeable improvement in the performance of coarse-grained classification. Adding specific alignment and feature enhancement strategies results in only slight improvements, indicating that the network is already capable of effectively discerning features under coarse classification conditions.

\textbf{Hierarchical Guidance Efficiently Enhances Fine-Grained WSI Classification.} Moreover, we compared methods relying on instance-level annotations to reveal the efficiency of our proposed framework in the task of fine-grained WSI classification of cervical cancer. We utilized the instance-level annotation of our CCC dataset, which containing 18,314 ROIs with 41,402 annotations in 5,332 WSIs in the training set. Following the original works of \cite{lin2021dual, cheng2021robust,zhang2022whole}, we reproduce these methods and present the results in Table \ref{ablation_4}. 

\begin{table}[h]
	\centering
	\caption{Comparison for fine-grained cervical cancer classification task, with the best results highlighted in bold.}
	\resizebox{0.49\textwidth}{!}{
		\begin{tabular}{c|ccccc}
			\toprule
			\textbf{Methods} & \textbf{Accuracy} & \textbf{Specificity} & \textbf{Sensitivity} & \textbf{F1} & \textbf{AUC}  \\
			\midrule
			ABMIL \cite{ilse2018attention} & 76.27 & 88.67 & 26.11 & 26.28 & 87.26 \\
			DSMIL (ImageNet Feature)\cite{li2021dual} & 75.95 & 88.19 & 24.52 & 24.43 & 85.51 \\
			DSMIL (Original Feature) \cite{li2021dual} & 76.42 & 91.03 & 24.12 & 24.07 & 87.92 \\
			Lin et al. \cite{lin2021dual} & 75.12 & 87.01 & 25.22 & 25.18 & 86.85 \\
			Cheng et al. \cite{cheng2021robust} & 79.20 & 91.12 & 35.30 & 36.25 & 90.03 \\
			Zhang et al. \cite{zhang2022whole} & 80.27 & 93.48 & \textbf{41.05} & \textbf{44.37} & \textbf{91.34} \\
			HMIL (Ours) & \textbf{80.39} & \textbf{93.60} & 40.84 & 44.13 & 91.16 \\
			\bottomrule
	\end{tabular}}
	\label{ablation_4}
\end{table}

From the results, we observed that our method has comparable performance to methods that rely on instance-level annotations, underscoring its efficiency. Specifically, Lin's method, equipped with a simple VGG-based detector and rule-based aggregation methods, even falls short of ABMIL, which does not rely on instance-level annotations. This indicates that reliance on instance-level annotations does not necessarily guarantee superior performance. On the other hand, Cheng and Zhang's methods both utilize several instance-level models to refine the results from the initial detector and achieve better performance. However, these methods come with the significant drawback of requiring instance annotations in various forms, which are time-consuming and labor-intensive to gather. DSMIL, while utilizing pretrained features, shows limited performance improvement and reduced sensitivity, indicating that pretraining encoders on a relatively small dataset may not be sufficient. Furthermore, when it directly leverages features extracted from the ImageNet-pretrained encoder, performance drops significantly, indicating that the non-local aggregator may not be ideally suited for fine-grained WSI classification tasks. In contrast, our HMIL framework comprehensively integrates hierarchical guidance into the MIL framework and achieves state-of-the-art performance, which underscores the technical path of reducing reliance on instance-level annotations to improve fine-grained WSI classification performance.

%% file: Sections/5_Conclusion.tex
\section{Discussion and Conclusion}
\label{sec:conclusion}
Our experiments demonstrate the efficacy of hierarchical alignment in enhancing both fine-grained and coarse-grained WSI classification through class-wise attention learning. However, the limitations in fine-grained classification sensitivity necessitate further investigation. At the feature extraction stage, recent pathology foundation models which pretrained on extensive pathological datasets \cite{chen2024towards, ma2024towards, mstar}, offer potential for more representative feature extraction and semantic interpretation. Integrating these models into our hierarchical framework or implementing hierarchical pretraining techniques \cite{yazdani2024helm} may potentially improve the current classification performance. At feature aggregation stage, state-space model-based approaches have exhibited encouraging results in MIL applications through their superior sequence modeling capabilities. Future research should explore incorporating hierarchical guidance into these architectures \cite{xu2024survey, fillioux2023structured, yang2024mambamil}. Moreover, addressing domain bias arising from imaging artifacts and tissue variations through domain adaptation and debiasing techniques is crucial for model robustness and generalizability across heterogeneous imaging modalities \cite{luo2024medical}. These insights underscore the synergistic importance of hierarchical guidance and domain adaptation in advancing WSI classification methodology.

In this work, we introduced the HMIL framework, an approach that leverages label hierarchy into the MIL framework to address the fine-grained WSI classification task. HMIL comprehensively aligns features from the instance to the bag level. The framework further incorporates dynamic weighting and supervised contrastive learning, which refine slide-level representations, resulting in improved classification outcomes. Crucially, HMIL eliminates the need for extensive instance-level annotations. It demonstrates robust performance across various WSI datasets from different imaging modalities, including the publicly available histology datasets BRACS and PANDA, as well as our extensive cervical cytology WSI dataset CCC, underscoring its effectiveness and adaptability.